\documentclass[lettersize,journal]{IEEEtran}
\usepackage{amsmath,amsfonts}
\usepackage{algorithmic}
\usepackage{algorithm}
\usepackage{array}
\usepackage[caption=false,font=normalsize,labelfont=sf,textfont=sf]{subfig}
\usepackage{textcomp}
\usepackage{stfloats}
\usepackage{url}
\usepackage{verbatim}
\usepackage{graphicx}
\usepackage{cite}
\usepackage{enumitem}
\usepackage{textcomp}
\usepackage{amsmath}
\usepackage{threeparttable}
\usepackage{booktabs}
\usepackage{tabularx}
\usepackage{xcolor}

\begin{document}

\title{ParkingTransformer: LLM-Enhanced End-to-End Trajectory Planning for Autonomous Parking}

\author{Hauteng Wu, Xu Li*, Dong Kong, Zihang Wang, Xieyuanli Chen, Benwu Wang, Wenkai Zhu
\thanks{Huateng Wu, Xu Li*, Zihang Wang, Benwu Wang and Wenkai Zhu are with the School of Instrument Science and Engineering, Southeast University, Nanjing 210096, China, e-mail:{\tt\small 101010791@seu.edu.cn}, * is the corresponding author.}
\thanks{Dong Kong is with the School of Electronic and Information Engineering, Tongji University, Shanghai, 201804, China, and also with the College of Transportation, Shandong University of Science and Technology, Qingdao 266590, China, e-mail: {\tt\small kongd\_6696@163.com}. }

\thanks{Xieyuanli Chen is currently an Associate Professor with the National University of Defense Technology. Dr. Chen was also the Associate Editor for IEEE TRO, RA-L, ICRA, and IROS, e-mail: {\tt\small chenxieyuanli@hotmail.com}. }

}

\markboth{}%
{Shell \MakeLowercase{\textit{et al.}}: A Sample Article Using IEEEtran.cls for IEEE Journals}


\maketitle

\begin{abstract}
End-to-end autonomous parking has emerged as a critical task within the realm of autonomous driving. However, existing methods suffer from black-box characteristics, lacking high-level semantic understanding and interpretability, which impedes the realization of seamless long-distance autonomous parking from the road to the target spot. To address these limitations, we propose ParkingTransformer, a novel framework that leverages multi-view perception and the scene understanding capability of Large Language Models (LLMs). By combining trajectory queries with LLMs implicit state features, our method interacts directly with historical information and raw sensor data to output planning trajectories, eliminating the need for dense Bird's-View (BEV) representations. To compensate for the inadequate spatial reasoning ability of LLMs, we introduce 3D positional encoding to explicitly inject spatial geometric awareness. Furthermore, a fixed-window streaming mechanism is designed for historical information processing, significantly improving long-term temporal processing efficiency and inference speed. Additionally, a coarse-to-fine decoding strategy is employed to progressively enhance trajectory precision. Extensive closed-loop experiments are conducted on the CARLA simulator and real-world vehicle platforms. The results demonstrate that our method achieves a driving score of 61.32 in CARLA simulator and an average success rate of 88.70\% in real-world experiments, validating the feasibility and effectiveness of the proposed algorithms. 
\end{abstract}

\begin{IEEEkeywords}
Autonomous Parking, End-to-End, Large Language Model, Trajectory Planning.
\end{IEEEkeywords}

\section{Introduction}
\IEEEPARstart{A}{utonomous} driving technology has emerged as a pivotal research domain in both academia and industry, owing to its potential to revolutionize transportation systems\cite{r1}.  Within this domain, autonomous parking systems serve as a critical means to achieve precise vehicle maneuvering\cite{r2}. Current parking technologies are predominantly proximity-based, with an operational range typically confined to the immediate vicinity of the parking spot (within 10 meters)\cite{r3}. However, a more pragmatic requirement is long-range autonomous parking, which entails a seamless, end-to-end autonomous transition from the road network (or parking lot entrance) to the target parking space\cite{r4}, usually covering distances of 50 to 300 meters. Research in the domain of long-distance autonomous parking remains scarce, and existing approaches predominantly adopt rule-based cascaded architectures comprising environmental perception, map construction, and planning and control modules\cite{r5}. Although such architectures simplify system design to some extent, the absence of end-to-end learning capabilities inevitably leads to information compression and error accumulation across modules, rendering the system highly susceptible to failures during long-range parking operations.

In recent years, end-to-end data-driven paradigms have offered a promising avenue to address complex parking tasks. The core advantage lies in integrating disparate modules into a unified neural network, thus eliminating reliance on intricate hand-crafted rules while exhibiting
\begin{figure}[!t]
\centering
\includegraphics[width=0.5\textwidth]{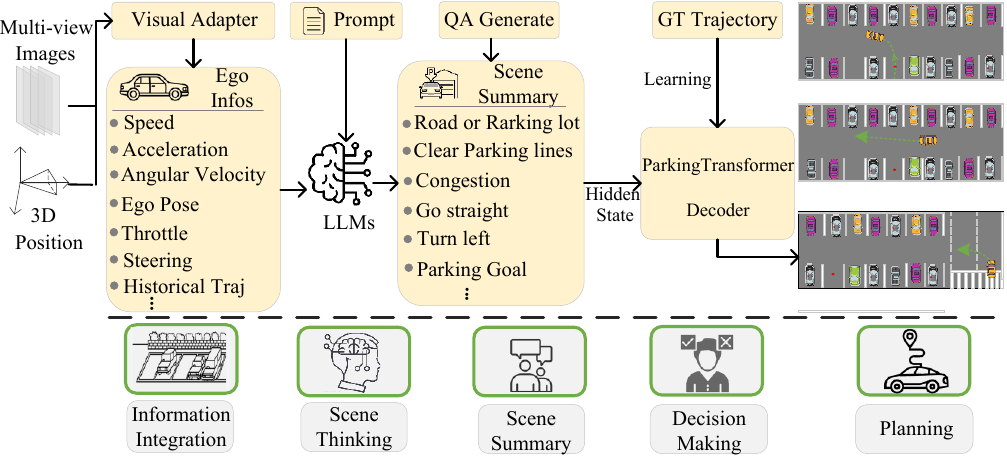}
\caption{We propose ParkingTransformer, a framework that leverages the powerful scene understanding capability of LLMs to transform multi-view visual features and ego-vehicle states into implicit states containing high-level semantics (e.g., scene summary, decision making). The model is capable of human-like reasoning and decision-making in long-distance parking scenarios, directly outputting trajectories.}
\label{fig_1}
\end{figure}
 scalability that improves progressively with increasing data volume\cite{r6}. Despite the demonstrated benefits of such methods in perception and planning, existing approaches suffer from at least three fundamental limitations. First, the construction of dense BEV features incurs substantial computational overhead. As shown in Fig. 2 (a), to address spatiotemporal complexity, methods such as LSS\cite{r7} and BEVFormer\cite{r8} rely on dense BEV feature representations, as perception range increases, computational costs escalate dramatically. These approaches often overlook the inherent sparsity of real-world scenes\cite{r9}, severely constraining inference speed. Second, there is a lack of comprehension capabilities for complex scenarios \cite{r9}, making it difficult to meet the precise motion control requirements of parking tasks. Although methods like ParkingE2E \cite{r10} adopt an end-to-end paradigm, their performance in long-distance parking is suboptimal due to the inability to generate descriptive representations of the parking scene. Furthermore, the opacity of decision-making logic precludes human interpretability, resulting in inadequate explainability when accidents occur. Third, as shown in Fig. 2 (b), recently, certain studies\cite{r11,r12,r13} have attempted to introduce LLMs datasets, leveraging LLMs-generated scene descriptions to enhance system explainability. However, LLMs inherently possess limited 3D spatial reasoning capabilities; directly utilizing their output hidden states for trajectory planning falls short of satisfying the requirements for fine-grained motion control in parking scenarios. Therefore, despite the significant potential demonstrated by current end-to-end frameworks, their deployment in long-distance parking scenarios remains severely constrained by three fundamental limitations: the prohibitive computational overhead associated with dense BEV representations; the inadequate comprehension of parking-specific scene semantics; and the insufficient spatial-geometric reasoning capabilities even when scene understanding is achieved, ultimately resulting in imprecise trajectory prediction.

\begin{figure}[!t]
\centering
\includegraphics[width=0.5\textwidth]{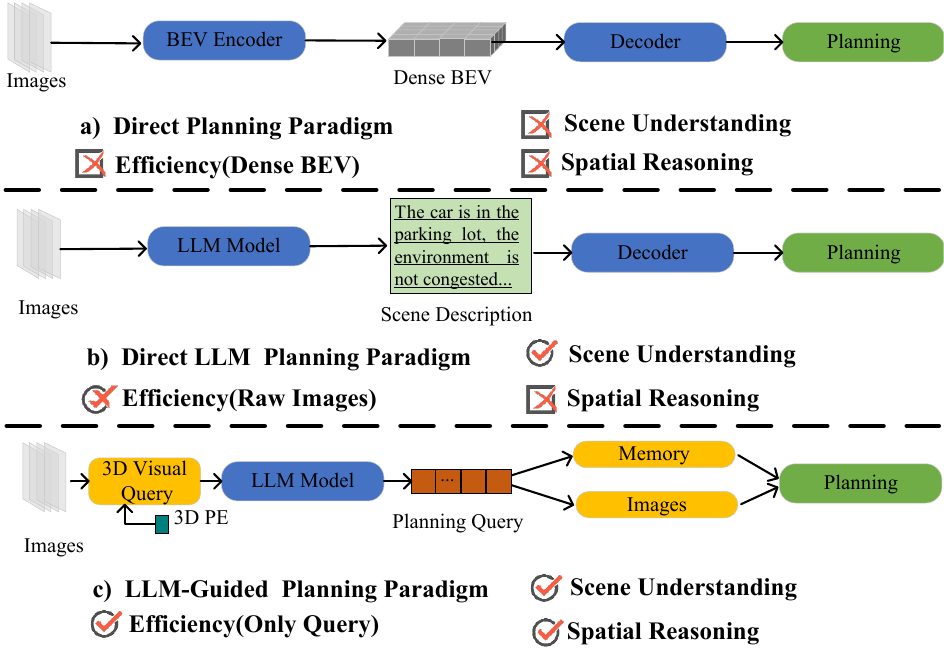}
\caption{End-to-End Autonomous Parking Paradigm Comparison. (a) Limited by dense BEV, leading to extensive redundancy. (b) Despite possessing scene understanding capabilities, direct LLM processing of raw images results in high computational costs and is restricted by limited spatial reasoning. (c) Transforms images into a set of queries interacting directly with memory and raw images, avoiding the generation of dense BEV features.}
\label{fig_1}
\end{figure}

To address these limitations, this paper proposes ParkingTransformer, a novel end-to-end autonomous parking framework that innovatively integrates multi-view visual perception with the semantic reasoning capabilities of LLMs. Specifically, we first leverage the reasoning capabilities of LLMs to generate a deep understanding and description of the scene. Subsequently, trajectory queries are fused with LLMs implicit state features; temporal cross-attention (TCA) is employed to extract historical information, while cross-attention with raw sensor data captures environmental cues; ultimately enabling direct trajectory generation. This design eliminates the necessity for dense BEV intermediate representations. To compensate for the inherent spatial reasoning deficiencies of LLMs, we introduce 3D positional encoding to explicitly inject spatial geometric awareness. Furthermore, to enhance long-term temporal processing efficiency, we devise a fixed-window streaming mechanism that significantly improves inference speed and computational efficiency. At the decoder, a coarse-to-fine strategy is adopted to progressively refine trajectory precision. The contributions of this paper are summarized as follows:

\begin{enumerate}[
    leftmargin=3em,     
    labelwidth=2em,     
    label=\arabic*),    
    align=right,        
    itemsep=1ex         
]
    \item We present a novel llm-enhanced framework for long-range autonomous parking. Our approach integrates multi-view visual perception and the semantic-spatial reasoning of LLMs into a unified end-to-end Transformer architecture.

    \item Trajectory queries are integrated with the implicit state features of the LLMs, interacting directly with historical data and raw sensor inputs via cross-attention. This allows for direct trajectory generation without relying on dense BEV intermediate representations.

    \item We leverage the reasoning capabilities of LLMs to generate scene understanding and descriptions, significantly enhancing the interpretability of the end-to-end framework. Concurrently, we introduce explicit 3D positional encoding to inject spatial geometric awareness, effectively compensating for the inherent limitations of LLMs in spatial reasoning.

    \item We devise a fixed-window streaming mechanism to substantially improve the efficiency and speed of processing historical information. Furthermore, a coarse-to-fine strategy is adopted in the decoder to further refine trajectory precision.

\end{enumerate}

\section{Related Work}

\subsection{End-to-End Autonomous Driving and  Parking}

In contrast to traditional autonomous driving frameworks governed by hand-crafted rules, the end-to-end paradigm employs neural networks to directly map raw sensor inputs to control commands or planning trajectories. This architecture effectively circumvents information loss and error propagation inherent in modular systems. The genesis of end-to-end driving can be traced back to Pomerleau \cite{r14}, with early research predominantly focusing on on-road scenarios. Modern approaches largely rely on imitation learning \cite{r15} to distill human-like driving policies from expert demonstrations. Representative works include CIL \cite{r16}, which pioneered a streamlined architecture mapping images directly to control signals, and CILRS \cite{r17}, which mitigated causal confusion during training through speed prediction as an auxiliary task. To handle the sequential nature of trajectory planning, GRU-based methods \cite{r18} have been widely adopted for auto-regressive prediction. Frameworks such as Transfuser \cite{r19}, Transfuser++ \cite{r20}, and NEAT \cite{r21} typically utilize encoder-extracted features for trajectory regression in the decoder stage. Recent advancements have focused on enhancing prediction fidelity and architectural efficiency. ThinkTwice \cite{r22} introduced a scalable coarse-to-fine decoder paradigm, refining initial trajectory estimates with detailed observational features to achieve higher precision. UniAD \cite{r1} pioneered a unified transformer architecture, integrating perception, prediction, and planning via query-based mechanisms to facilitate cross-module interaction and joint optimization. Furthermore, to address the computational burden of dense BEV representations, VAD \cite{r23} proposed vectorized scene modeling, explicitly outputting vectorized dynamic objects and static map elements.

Compared with on-road driving scenarios, end-to-end autonomous parking has received relatively limited research attention. Rathour et al. \cite{r24} attempted to fuse visual and dead-reckoning information to regress parking trajectories through neural networks. Subsequently, Li et al. \cite{r25} proposed a specialized algorithm for vertical parking slots, utilizing a CNN (Convolutional Neural Network) to output steering and throttle commands. However, this approach remains partially tethered to geometric constraints, thereby precluding fully end-to-end learning. To capture temporal dynamics, ParkPredict \cite{r26} employed recurrent neural networks (RNNs) and long short-term memory (LSTM) units to jointly predict trajectory points and vehicle intentions. Building upon this, ParkPredict++ \cite{r27} adopted a transformer architecture, leveraging self-attention mechanisms to model long-range dependencies between the ego vehicle and its environment, thus generating multi-modal probabilistic trajectories. Yang et al. \cite{r28} first validated the feasibility of end-to-end parking in the CARLA simulator, directly mapping surround-view images to control signals through imitation learning. ParkingE2E \cite{r10} extended this paradigm to real-world scenarios by incorporating BEV representations and target query mechanisms, utilizing an autoregressive transformer decoder to predict waypoints with an 87.8\% success rate across four real-world parking garages. In summary, existing approaches either suffer from a persistent reliance on geometric rules or lack the interpretability of intermediate decision processes, making the model's decisions opaque. Furthermore, constrained by dense BEV intermediate representations, these methods incur substantial computational overhead and lack deep semantic understanding of complex scenes. Consequently, they struggle to meet the stringent requirements of long-distance autonomous parking tasks from the road network to the target slot.

\begin{figure*}[htbp]
\centering
\includegraphics[width=1.0\textwidth]{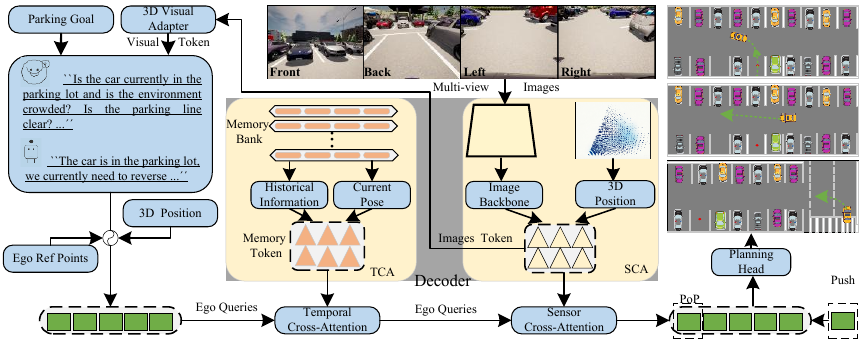}
\caption{Overall Framework. LLMs generate implicit state representations that encapsulate the environmental context, ego-vehicle dynamics, and the target parking slot. These latent features are subsequently processed by the TCA (Temporal Cross-Attention) and the SCA (Sensor Cross-Attention) to yield predicted trajectories. For historical information processing, we implement a fixed-window streaming mechanism governed by a FIFO (First-In-First-Out) strategy. Notably, the SCA eschews dense BEV representations, instead interacting directly with raw sensor data and employing a coarse-to-fine decoding paradigm to progressively refine trajectory predictions.
.}
\label{fig2}
\end{figure*}

\subsection{LLMs in Autonomous Driving}

LLMs endowed with powerful semantic understanding and scene description capabilities have recently been introduced into the autonomous driving domain to enhance system interpretability. LMDrive\cite{r29} proposed a framework that leverages LLMs for closed-loop end-to-end driving, transforming vehicle state information (e.g., speed, acceleration) and environmental perception cues (e.g., bounding boxes, semantic labels) into structured natural language sequences, through which the LLMs directly generates control command sequences encompassing steering, throttle, and brake. To bridge the gap between pure visual perception and language model reasoning. DriveVLM \cite{r30} fused visual features with linguistic instructions, feeding them into a unified multi-modal large model. DriveGPT4\cite{r31} presented an interpretable end-to-end autonomous driving architecture that not only outputs vehicle control signals but also generates detailed reasoning processes and decision rationales, substantially enhancing system explainability. OmniDrive \cite{r32} constructed an LLMs-agent framework that first builds an environmental world model through a 3D perception module, then employs the LLMs for complex scene understanding and interactive reasoning, ultimately generating globally consistent motion trajectories compliant with traffic rules. However, existing LLMs-based approaches predominantly focus on high-level decision-making for on-road driving scenarios. Their hidden states lack precise geometric and spatial encoding, resulting in insufficient spatial reasoning capabilities that fall short of satisfying the centimeter-level trajectory precision required by parking tasks. How to effectively integrate the semantic understanding advantage of LLMs with the spatial geometric information of visual perception, to enhance spatial reasoning ability while achieving precise trajectory generation, remains an under-explored direction.

\section{Methodology}
This paper proposes an autonomous parking framework that integrates multi-view visual perception with the semantic reasoning capabilities of LLMs. Figure 3 illustrates our approach, which combines trajectory queries with the implicit state features of LLMs. By employing two cross-attention mechanisms to extract historical and raw sensor information respectively, the model directly outputs the planned trajectory. Our model eliminates the reliance on dense BEV intermediate representations. To compensate for the inherent limitations of LLMs in spatial reasoning, we introduce 3D positional encoding to explicitly inject spatial geometric perception. Furthermore, to enhance long-term temporal processing efficiency, we design a fixed-window-based streaming mechanism that significantly improves inference speed. Finally, a coarse-to-fine strategy is adopted at the decoder to further elevate the precision of trajectory planning.

\begin{figure*}[htbp]
\centering
\includegraphics[width=1.0\textwidth]{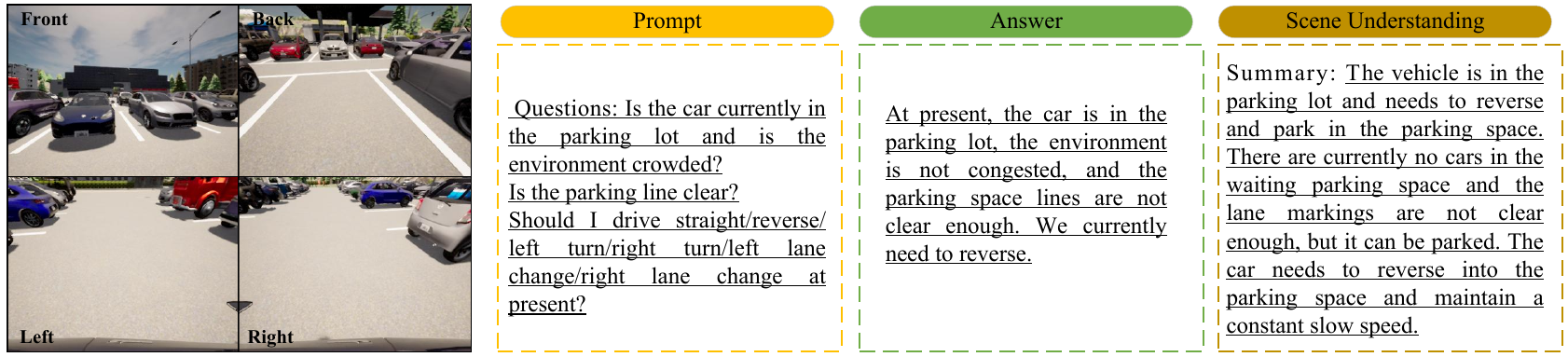}
\caption{Scene understanding and implicit state generation.}
\label{fig3}
\end{figure*}

\begin{figure}[!t]
\centering
\includegraphics[width=0.4\textwidth]{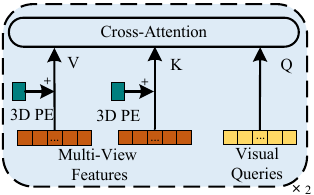}
\caption{3D Visual Adapter.}
\label{fig_1}
\end{figure}

\subsection{Problem Definition}

Given four camera images captured at time $t$ , $ I_t=\{I_t^{front}, I_t^{back}, I_t^{left}, I_t^{right}\}$, the motion state of the autonomous vehicle $S_{ego}=\{S_{acc}, S_{gyr}, S_v, w, l, Steer, Pedal, His\__{traj}, P_{ego}, P_{goal}\}$, including vehicle acceleration, angular velocity, velocity, vehicle width, vehicle height, steering wheel angle, pedal, historical trajectory, ego vehicle global pose $P_{ego}$, and global pose of the target parking slot $P_{goal}$, Implicit state features $T$ output by the LLMs model, the algorithm goal is to predict the autonomous vehicle trajectory points at future $T_{pred}$ times, $\tau=\{x_t, y_t\}_{t=1}^{T_{pred}}$, $T_{pred} = 30$.

\subsection{Multi-Perspective Feature Embedding and LLMs Scene Understanding}
Given the current 4-channel surround-view images, we utilize ResNet-50 \cite{r33} as the backbone to extract multi-scale image features. These features are subsequently fused via a Feature Pyramid Network (FPN) \cite{r34} to yield visual feature maps $F_t \in \mathbb R^{B\times N \times C\times H\times W}$ with unified dimensions. We adopt Qwen2.5\cite{r35} as our LLMs. Since the features derived from the backbone and neck are not directly compatible with the LLMs, a visual feature adapter is required. As illustrated in Fig. 5, we draw inspiration from Q-Former \cite{r36} as the visual adapter. The visual features are projected into queries via a multi-layer Perceptron (MLP), denoted as $Queries  \in \mathbb R^{B\times Q\times D}$, where Q represents the number of queries for the surround-view cameras. To balance performance and efficiency, we assign 8 queries per camera, resulting in a total of Q =32. Let L denote the text embedding length of the LLM (defaulting to 896). To compensate for the limited spatial reasoning capabilities of LLMs, inspired by PETR \cite{r37}, we incorporate 3D positional encoding into the image features. The queries then undergo two layers of cross-attention to obtain the visual features required by the LLMs. As shown in Figure 4. We design a series of critical prompts to direct the LLM's attention to key regions within the scene. The LLMs ultimately output responses and scene descriptions based on these prompts. The final layer's $hidden\_state  \in \mathbb R^{B\times 1\times L}$ is fused with the trajectory query via an MLP to generate the initial trajectory $query  \in \mathbb R^{B\times 1\times L}$, where $L$ represents the feature dimension.  The initial trajectory queries encode high-level semantic information of the scene, along with ego-vehicle states and parking destination details, such as whether the vehicle is inside a parking lot, whether the environment is congested, and whether the current maneuver requires moving forward, turning left/right, changing lanes left/right, reversing, and the target parking spot, thereby providing implicit conditioning for subsequent trajectory generation.

\subsection{Memory Token and Fixed-window Streaming Mechanism}

Autonomous parking tasks necessitate not only current scene semantics, ego-vehicle states, and target slot information, but also historical temporal context, which encompasses prior knowledge of the environment and the ego vehicle's historical trajectory. We integrate historical information through a TCA mechanism. First, since the coordinate frames differ across historical timesteps, we transform the ego vehicle's poses from all historical moments into the current coordinate frame. Based on the current pose, we compute the 3D positional encodings for the current timestep. Inspired by PETR\cite{r37}, these positional encodings incorporate both spatial alignment and feature alignment. Leveraging historical and current scene information, along with the future parking target and ego state, we compute the initial parking trajectory. Specifically, the target parking slot information includes the global position, orientation, width, and height. Subsequently, the current ego query and its positional encodings are fed into the TCA mechanism to fuse historical information, where the query is the current ego query, while the key and value are the historical ego queries. In this manner, the ego query can integrate information from both history and the present, including environmental context, historical trajectories, ego state and target slot information. Maintaining a complete historical sequence incurs substantial computation and storage overhead. Therefore, this paper proposes a streaming memory mechanism with a fixed window. Specifically, we maintain only the historical memory features $M\_hist\in \mathbb R^{K \times D}$ for the most recent K frames, to balance performance and efficiency, $K$ is set to an empirical value of 10, updated in a queue format. When a new frame arrives, the feature of the oldest frame is dequeued, and the current frame is enqueued. This mechanism retains the necessary temporal context while controlling computational complexity. By maintaining historical data within a fixed window via a First-In-First-Out (FIFO) queue, we reduce computational overhead while effectively fusing historical information.
\begin{equation}
\label{deqn_ex1a}
PE_t = MPL(T_{his}^tPose_{his})
\end{equation}

\begin{equation}
\label{deqn_ex2a}
\begin{split}
H_t = \text{TCA-Attention} (&Q = H_{ego} + PE_t, \\
                            &K = H_{ego}^{his} + PE_{his}, \\
                            &V = H_{his}^{ego})
\end{split}
\end{equation}

\begin{figure}[!t]
\centering
\includegraphics[width=0.5\textwidth]{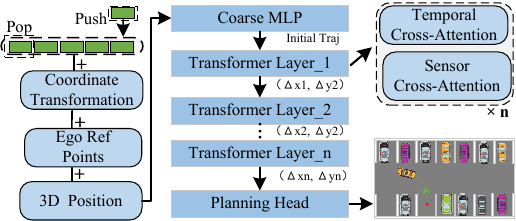}
\caption{Coarse-to-Fine Decoder.}
\label{fig_1}
\end{figure}

\subsection{Sensor Token and Coarse-to-fine Mechanism}
The multi-view images are processed by ResNet-50 \cite{r33} to extract image features $F_t \in \mathbb R^{B\times N \times C \times H\times W}$, where B denotes the batch size, $N$ is the number of cameras, $C$ represents the channel dimension, $H$ and $W$ correspond to the height and width of the feature maps. After incorporating 3D positional encoding, these features serve as the keys and values for the SCA (Sensor Cross-Attention) mechanism. For each pixel, we pre-define $D$ depth candidates (set to $D$ =64). Since parking scenarios primarily focus on nearby objects, we adopt the Logarithmic Interval Discretization\cite{r38} (LID) strategy, which employs dense sampling for near-field depths and sparse sampling for far-field depths, as shown in Equation 4. By applying 3D positional encoding to the image features, we enhance the perception of spatial geometry and semantics. Notably, to maintain consistency with the input features of the LLMs, the 3D positional encoding used here is aligned with that of the LLMs module. Furthermore, we employ a coarse-to-fine mechanism, as shown in Figure 6. Initially, a coarse parking trajectory is computed based on historical and current scene information, future parking targets, and the ego-vehicle's state. This trajectory is then processed through a multi-layer transformer decoder. In this architecture, the trajectory output from each decoder layer serves as a refinement residual, which is added to the coarse trajectory before being fed into the subsequent layer. This progressive refinement mechanism allows the network to first capture the global trend of the trajectory and subsequently adjust local details, effectively improving adaptability to complex parking scenarios.

\begin{equation}
\label{deqn_ex3a}
\begin{split}
H_t = \text{SCA-Attention} (&Q = H_{ego} + PE_t, \\
                            &K = H_{ego}^{img} + PE_{t}, \\
                            &V = H_{ego}^{img})
\end{split}
\end{equation}

\begin{equation}
\label{deqn_ex4a}
\begin{split}
d_i = dmin + \frac{d_{max}-d_{min}}{N-1} \cdot(\frac{e^\frac{i}{N-1}\cdot ln(\frac{d_{max}}{d_{min}}+1)-1}{\frac{d_{max}}{d_{min}}})
\end{split}
\end{equation}

\subsection{Trajectory Planning Head and Loss Function}

After passing through the 12 decoder layers, the ego query is fed into the trajectory planning head at each layer to output trajectories, where the trajectory planning head is implemented as an MLP. This paper adopts an end-to-end training paradigm. The total loss function consists of a weighted trajectory regression loss and an LLMs loss. Specifically, the loss is computed at each decoder layer. The trajectory planning loss employs the L1 loss $\mathcal{L}_{reg}$, while the text output from the LLMs is supervised by the cross-entropy loss $\mathcal{L}_{ce}$. Considering that the magnitude of the LLMs loss is substantially larger than that of the trajectory loss, we apply weighted balancing to these two loss terms, $\lambda = 1.5$. Due to the nature of long-distance parking, short trajectories in the final parking stage are relatively scarce. Consequently, the learning process tends to be biased towards medium-to-long trajectories. To address this, we apply a weighted loss to the short trajectories, as shown in Equation 6. Specifically, weights of 20.0 and 5.0 are assigned to trajectories shorter than 0.5 m and 2 m, respectively.

\begin{equation}
\label{deqn_ex5a}
\begin{split}
\mathcal{L}_{total} = \lambda \mathcal{L}_{reg} + \mathcal{L}_{ce}
\end{split}
\end{equation}

\begin{equation}
\label{deqn_ex6a}
\begin{split}
w_i = \begin{cases}
    20.0, & \text{if } \tau _{gt}^i < 0.5\text{m} \\
    5.0,  & \text{if } \tau _{gt}^i < 2.0\text{m} \\
    1.0,  & \text{otherwise}
\end{cases}
\end{split}
\end{equation}

\section{Experiments}
\subsection{Implementation Details}

We conducted both simulation experiments and real-world vehicle experiments. For the simulation experiments, we utilized the CARLA 0.9.11 simulator for data collection and closed-loop evaluation. The simulation scenario was set in the parking lot of CARLA Town04, which comprises 64 parking spaces. For each trajectory, the ego vehicle was randomly initialized near the parking lot, navigated into the parking lot, and parked into the target parking space. A total of 200 clips were collected, with 150 clips allocated for training and 50 clips for testing. For the real-world vehicle experiments, we conducted real-vehicle experiments in an outdoor parking lot, which contains 60 parking spaces. Following the same protocol as the simulation experiments, trajectories were collected starting from near the parking lot with the target parking space as the destination. A total of 100 clips were collected, with 90 clips used for training and 50 clips for testing. The simulated and real-world parking lots are shown in Figure 7.

The models were trained on two NVIDIA GeForce RTX 4090 for a total of 70 epochs, with a batch size of 6, a learning rate of 1e-5, a weight decay of 0.01, and a dropout rate of 0.1. The AdamW optimizer was employed, with a warm-up period of 500 steps. The historical queue length K in the TCA was set to 10. The training was implemented using the PyTorch framework, with a simulation training duration of 27 hours and a real-world vehicle experiment training duration of 18 hours.

\begin{figure}[!t]
\centering
\includegraphics[width=0.5\textwidth]{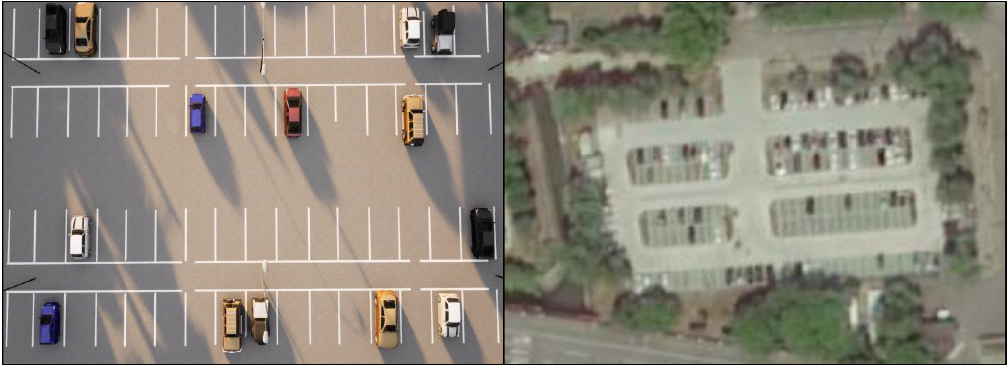}
\caption{
CARLA Town04 parking lot top view and outdoor parking lot top satellite view.}
\label{fig_7}
\end{figure}

\begin{figure}[!t]
\centering
\includegraphics[width=0.4\textwidth]{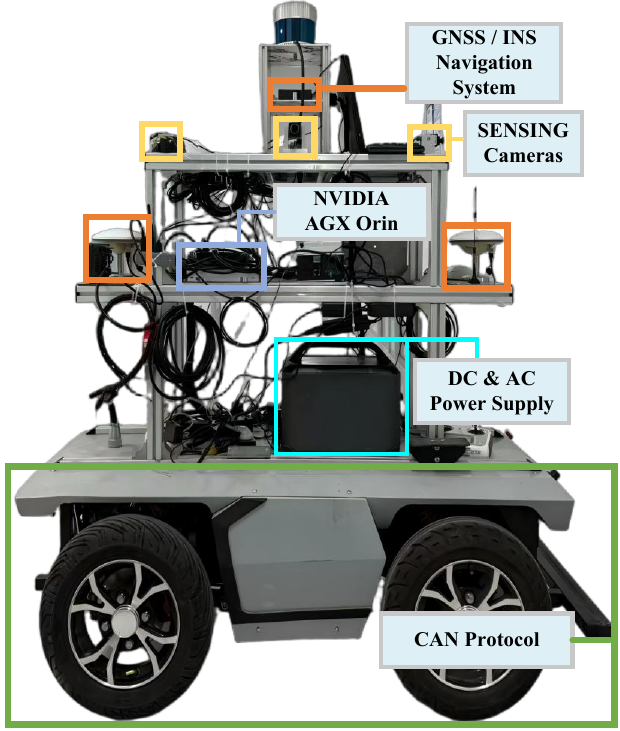}
\caption{
Autonomous Driving Vehicle platform. The mobile data acquisition platform integrates four SENSING RGB cameras, an integrated navigation system, a CAN bus interface for chassis communication and control, and an onboard computing unit dedicated to time-synchronized multi-modal data acquisition.}
\label{fig_8}
\end{figure}

\subsection{Data Collection}

Simulation Data Collection. We developed a data collection system based on the CARLA simulator. The collection scenario was set in the parking lot of CARLA Town04. The vehicle starting point was randomly generated at a nearby location outside the parking lot (either on the road or at the parking lot entrance), while the target parking space was predetermined and always unoccupied. The occupancy status of other parking spaces in the lot was randomly assigned. The distance between the starting point and the target space ranged from 10 to 200 meters. Each ground-truth (GT) trajectory was manually generated by an experienced driver using keyboard control. The vehicle was equipped with four RGB cameras (front, rear, left, and right). An IMU sensor provided angular velocity and acceleration data, while the CARLA simulator supplied the ego vehicle's velocity, global pose, throttle, and steering wheel angle. Trajectories were recorded only when the final parking distance to the target space was less than 0.3 m and the heading angle deviation was less than 1 degree, trajectories failing to meet these criteria were discarded. 

Real-world trajectories were collected at the outdoor parking lot using an experimental vehicle. The platform is equipped with four Sensing cameras for RGB image acquisition and a combined navigation system that provides angular velocity, acceleration, and global poses. Additionally, wheel speed sensors measure vehicle velocity, while the chassis outputs throttle position and front wheel steering angles via the CAN protocol. Five experienced drivers recorded all clips. Given the challenges associated with real-world maneuvering, trajectories satisfying the parking criteria with a final Euclidean distance of less than 0.6 m and a heading error of less than 8° were retained, while those failing to meet these thresholds were discarded. Figure 8 illustrates the mobile data acquisition platform.

\begin{table}[ht]
\centering
\begin{threeparttable}
\begin{minipage}{0.5\textwidth}
\caption{Comparison of Short-Range Parking Performance in CARLA (Distance $ \leq $ 20m) \label{tab:table1}}
\begin{tabular}{c|c c c}
\toprule
Methods                  & L2 Dis.(m)$\downarrow$  & Haus Dis.(m)$\downarrow$       & Four Diff.(m)$\downarrow$   \\
\midrule
TransFuer\cite{r19}      & 0.73                 & 1.32                            & 13.46                  \\
E2EParking\cite{r28}     & 0.12                 & 0.26                            & 2.32                   \\
ParkFormer\cite{r39}     & 0.31                 & 0.49                            & 5.71                   \\

Ours                     & \textbf{0.03}        &\textbf{0.08}               & \textbf{0.56}               \\
\bottomrule
\end{tabular}
\end{minipage}
\end{threeparttable}
\end{table}

\begin{table}[ht]
\centering
\begin{threeparttable}
\begin{minipage}{0.5\textwidth}
\caption{Comparison of Long-Range Parking Performance in CARLA (50 m$ \leq $ Distance $ \leq $ 300m)} \label{tab:table2}
\begin{tabular}{c|c c c}
\toprule
Methods                  & L2 Dis.(m)$\downarrow$  & Haus Dis.(m)$\downarrow$     & Four Diff.(m)$\downarrow$   \\
\midrule
TransFuer\cite{r19}      & \textbackslash     & \textbackslash                  & \textbackslash           \\
E2EParking\cite{r28}     & \textbackslash     & \textbackslash                  & \textbackslash           \\
ParkFormer\cite{r39}     & \textbackslash     & \textbackslash                  & \textbackslash           \\
Ours                     & \textbf{0.06}      &\textbf{0.17}                    & \textbf{0.86}            \\
\bottomrule
\end{tabular}
\end{minipage}
\end{threeparttable}
\end{table}

\begin{table}[ht]
\centering
\begin{threeparttable}
\begin{minipage}{0.5\textwidth}
\caption{Comparison of Short-Range Parking Performance in Real-World (Distance $ \leq $ 20m) \label{tab:table3}}
\begin{tabular}{c|c c c}
\toprule
Methods                  & L2 Dis.(m)$\downarrow$  & Haus Dis.(m)$\downarrow$    & Four Diff.(m)$\downarrow$   \\
\midrule
ParkingE2E\cite{r10}      & 0.25                 & 0.69                        & 10.42                  \\
Ours                     & \textbf{0.08}        &\textbf{0.21}                & \textbf{1.25}             \\
\bottomrule
\end{tabular}
\end{minipage}
\end{threeparttable}
\end{table}

\begin{table}[ht]
\centering
\begin{threeparttable}
\begin{minipage}{0.5\textwidth}
\caption{Comparison of Long-Range Parking Performance in Real-World (50 m$ \leq $ Distance $ \leq $ 300m)} \label{tab:table4}
\begin{tabular}{c|c c c}
\toprule
Methods                  & L2 Dis.(m)$\downarrow$  & Haus Dis.(m)$\downarrow$    & Four Diff.(m)$\downarrow$    \\
\midrule
ParkingE2E\cite{r10}     & \textbackslash       & \textbackslash               & \textbackslash               \\
Ours                     & \textbf{0.13}        &\textbf{0.31}                & \textbf{2.44}                 \\
\bottomrule
\end{tabular}
\end{minipage}
\end{threeparttable}
\end{table}

\begin{table*}[ht]
\centering
\begin{threeparttable}
\caption{ CARLA experiment of parking performance (Distance $ \leq $ 20m) \label{tab:table5}}
\begin{tabular}{c|c c c c c c c c}
\toprule
Methods                  & TSR(\%)$\uparrow$  & TFR(\%)$\downarrow$    & NTR(\%)$\downarrow$    & CR(\%)$\downarrow$    & APE(m)$\downarrow$     & AOE(m)$\downarrow$    & AOT(s)$\downarrow$     & Latency(ms)$\downarrow$\\
\midrule
TransFuer\cite{r19}       & 82.12              & 11.48                  & 3.73                  & 2.67                  & 0.52                     & 7.96                  & 17.6           & 48.6 \\
E2EParking\cite{r28}      & 90.31              & 4.22                   & 3.23                  & 2.24                  & 0.39                     & 1.81                  & \textbf{16.8}  & \textbf{40.0}     \\
ParkFormer\cite{r39}      & 85.72              & 9.77                   & 2.56                  & 1.95                  & 0.22                     & 1.09                  & 19.4           & 42.1          \\
Ours                      & \textbf{97.67}     & \textbf{1.19}          &\textbf{0.51}         & \textbf{0.63  }       & \textbf{0.03}             & \textbf{1.03}         & 21.2           & 198.3          \\    
\bottomrule
\end{tabular}
\end{threeparttable}
\end{table*}

\begin{table*}[ht]
\centering
\begin{threeparttable}
\caption{ CARLA experiment of parking performance (50 m$ \leq $ Distance $ \leq $ 300m)} \label{tab:table6}
\begin{tabular}{c|c c c c c c c c}
\toprule
Methods                  & TSR(\%)$\uparrow$  & TFR(\%)$\downarrow$    & NTR(\%)$\downarrow$    & CR(\%)$\downarrow$    & APE(m)$\downarrow$     & AOE(m)$\downarrow$    & AOT(s)$\downarrow$     & Latency(ms)$\downarrow$\\
\midrule
TransFuer\cite{r19}      & \textbackslash      & \textbackslash        & \textbackslash         & \textbackslash       & \textbackslash          & \textbackslash         & \textbackslash      &50.1  \\
E2EParking\cite{r28}     & \textbackslash      & \textbackslash        & \textbackslash         & \textbackslash       & \textbackslash          & \textbackslash         & \textbackslash      &\textbf{49.3}   \\
ParkFormer\cite{r39}     & \textbackslash      & \textbackslash        & \textbackslash         & \textbackslash       & \textbackslash          & \textbackslash         & \textbackslash      & 55.9  \\
Ours                     & \textbf{93.22}      & \textbf{3.89}          &\textbf{1.75}          & \textbf{1.14}        & \textbf{0.06}           & \textbf{1.83}          & \textbf{82.63}      & 201.7  \\    
\bottomrule
\end{tabular}
\end{threeparttable}
\end{table*}

\begin{table*}[ht]
\centering
\begin{threeparttable}
\caption{Real-world experiment of parking performance (Distance $ \leq $ 20m)\label{tab:table7}}
\begin{tabular}{c|c c c c c c c c}
\toprule
Methods                  & TSR(\%)$\uparrow$  & TFR(\%)$\downarrow$    & NTR(\%)$\downarrow$    & CR(\%)$\downarrow$    & APE(m)$\downarrow$     & AOE(m)$\downarrow$    & AOT(s)$\downarrow$  &Latency(ms)$\downarrow$ \\
\midrule

ParkingE2E\cite{r10}      & 89.24              & 5.63                  & 3.74                  & 1.42                  & 0.58                     & 3.33                 & \textbf{18.3}        & \textbf{52.7}  \\
Ours                      & \textbf{91.83}     & \textbf{4.32}          &\textbf{2.65}          & \textbf{1.20}         & \textbf{0.08}            & \textbf{1.66}        & 23.4                 & 221.9 \\    
\bottomrule
\end{tabular}
\end{threeparttable}
\end{table*}

\begin{table*}[ht]
\centering
\begin{threeparttable}
\caption{Real-world experiment of parking performance (50 m$ \leq $ Distance $ \leq $ 300m)}\label{tab:table8}
\begin{tabular}{c|c c c c c c c c}
\toprule
Methods                  & TSR(\%)$\uparrow$  & TFR(\%)$\downarrow$    & NTR(\%)$\downarrow$    & CR(\%)$\downarrow$    & APE(m)$\downarrow$     & AOE(m)$\downarrow$    & AOT(s)$\downarrow$  &Latency(ms)$\downarrow$ \\
\midrule

ParkingE2E\cite{r10}      & \textbackslash   & \textbackslash          & \textbackslash          & \textbackslash       & \textbackslash          & \textbackslash       & \textbackslash     & \textbf{57.5}     \\
Ours                      & \textbf{88.70}     & \textbf{6.13}          &\textbf{2.41}          & \textbf{2.76}         & \textbf{0.13}          & \textbf{3.92}         & \textbf{79.3}      & 231.2 \\    
\bottomrule
\end{tabular}
\end{threeparttable}
\end{table*}

\subsection{Quantitative Experiments}
According to SAE standards\cite{r3}, short-range parking is defined as within 50 meters, while long-range parking extends beyond 50 meters. Accordingly, we conduct experiments for both scenarios, with short-range parking performed at approximately 20 meters and long-range parking conducted at distances ranging from 50 to 300 meters. For both simulation and real-world vehicle experiments, we employ the L2 distance, Hausdorff  Distance (Haus. Dis.), and Fourier Descriptor Difference (Four. Diff.) for evaluation. Specifically, the L2 distance quantifies the average Euclidean distance between the predicted trajectory and the ground truth (GT). The Hausdorff distance, defined as the maximum of the minimum distances between two trajectories, quantifies the degree of matching between the prediction and the GT trajectory. Furthermore, the Fourier Descriptor Difference measures the discrepancy between trajectory shapes; a larger value indicates a greater divergence, as shown in Equation 7.

We select TransFuser\cite{r19}, E2EParking\cite{r28}, and ParkFormer\cite{r39} as baseline models. The CARLA simulation results for short-range and long-range parking are presented in Tables $\mathrm{I}$ and $\mathrm{II}$, while the real-world vehicle experimental results for short-range and long-range parking are reported in Tables $\mathrm{III}$ and $\mathrm{IV}$, owing to practical limitations in data labeling and auxiliary task integration, real-world comparisons are conducted exclusively against ParkingE2E\cite{r10}. As evidenced by both simulation and real-world experiments, ParkingTransformer achieves comparable performance to the aforementioned baselines in short-range parking scenarios, with marginal improvements over these methods. However, as the parking distance increases, the baseline methods fail in long-range parking scenarios due to their inadequate scene comprehension capabilities and the absence of historical information modeling and coarse-to-fine refinement mechanisms, rendering them incapable of handling the complexities associated with extended-distance parking maneuvers.

Furthermore, we employed the seven evaluation metrics from ParkingE2E\cite{r10} for a comprehensive assessment. Please refer to the Appendix for detailed definitions, with the corresponding results shown in Tables $\mathrm{V}$-$\mathrm{VIII}$. The results demonstrate that ParkingTransformer performs well across the aforementioned metrics. Although the inference speed is relatively lower, it remains acceptable for parking scenarios where vehicle speeds are typically below 15 km/h. Furthermore, while the compared methods exhibit advantages in inference speed, none of them incorporate LLMs, and this omission leads to deficiencies in scene understanding and spatial reasoning capabilities.

\begin{equation}
\label{deqn_ex7a}
\begin{split}
z_P(n) = x_P(n) + j \cdot y_P(n)  \\
z_G(n) = x_G(n) + j \cdot y_G(n)
\end{split}
\end{equation}

\begin{figure}[!t]
\centering
\includegraphics[width=0.5\textwidth]{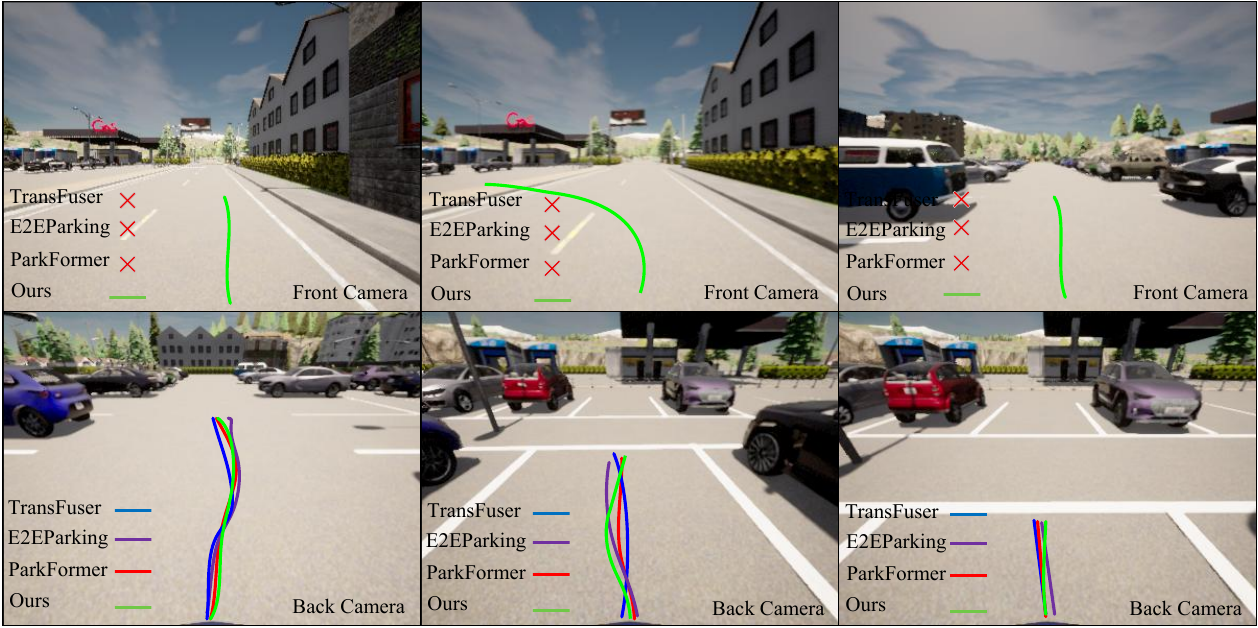}
\caption{
Visualization of autonomous parking trajectory planning in the CARLA simulation environment.}
\label{fig_9}
\end{figure}

\begin{figure}[!t]
\centering
\includegraphics[width=0.5\textwidth]{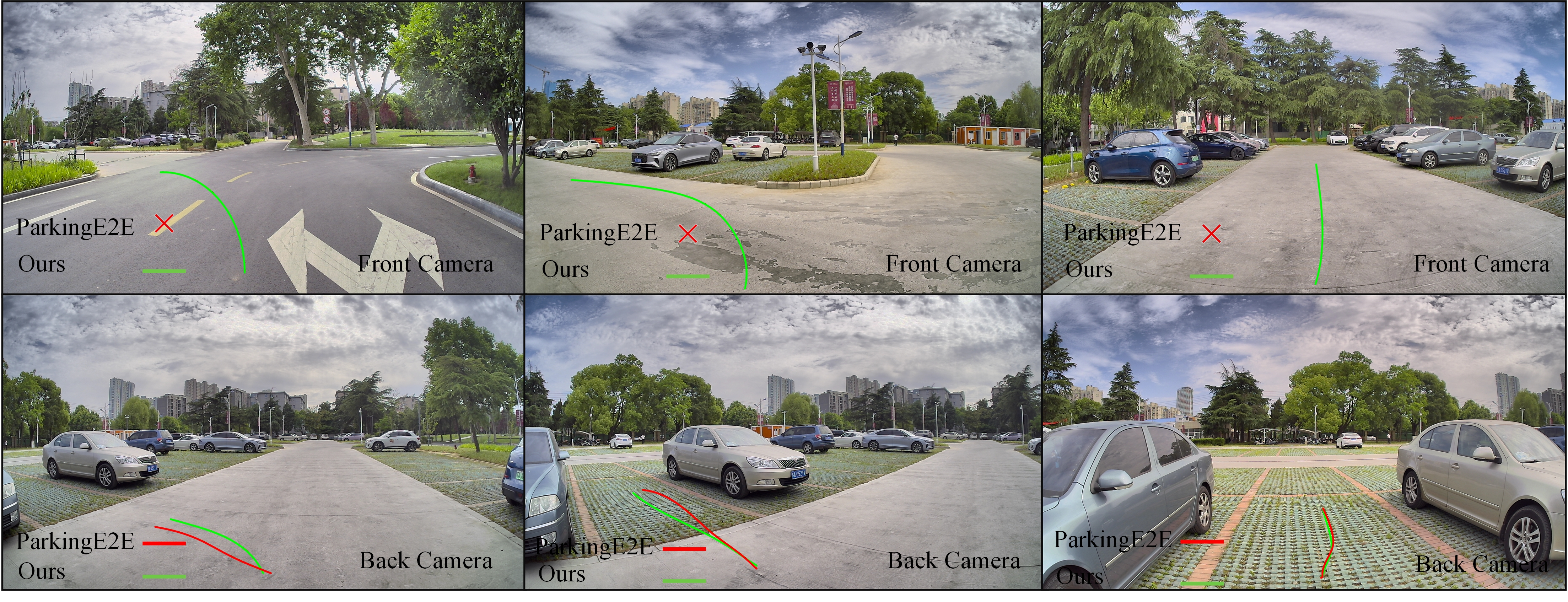}
\caption{
Visualization of autonomous parking trajectory planning in the real-world environment.}
\label{fig_10}
\end{figure}

\begin{equation}
\label{deqn_ex8a}
\begin{split}
F_p(k) = |\sum_{n=0}^{N-1}Z_p(n)e^{-j\frac{2\pi}{N}kn}| \\
F_G(k) = |\sum_{n=0}^{N-1}Z_G(n)e^{-j\frac{2\pi}{N}kn}|
\end{split}
\end{equation}

\begin{equation}
\label{deqn_ex9a}
\begin{split}
Four.Diff. = \sqrt{\sum_{k=0}^{K-1}(F_P(k)-F_G(k))^2}
\end{split}
\end{equation}

\subsection{Qualitative Experiment Visualization}
We conducted experiments in both the CARLA simulator and on real vehicles, which were similarly divided into short-trajectory and long-trajectory parking tasks. We performed short-distance experiments within a 20-meter range near the parking spaces (see the second row, rear-view camera, in Figures 9-10). For the long-trajectory experiments, which spanned 50 to 300 meters, the routes began at the parking lot's nearby roads or entrances, as depicted in the first two rows (front-view and rear-view cameras) of Figures 9-10, the results are consistent with our previous quantitative findings. In short-distance parking experiments, ParkingTransformer performs on par with the baseline methods. However, in long-distance experiments, the baseline models almost entirely collapse. As the distance increases, these baseline methods fail to handle long-range parking due to their lack of scene understanding, historical information, and coarse-to-fine mechanisms, ultimately leading to task failure. In contrast, ParkingTransformer is capable of planning a feasible trajectory from the road near the parking lot to the target parking slot, successfully executing the complete maneuver of entering the lot, navigating to the vicinity of the target slot, and finally reversing into the space.

\subsection{Ablation Experiment}
We designed three ablation studies to investigate the specific contributions of individual modules. As illustrated in Table $\mathrm{IX}$, both the TCA and SCA modules contribute to parking performance, with the SCA playing the most critical role. Because the vehicle requires environmental awareness for safe navigation, without the SCA, the agent effectively operates without visual perception. Table $\mathrm{X}$ demonstrates the significant impact of the LLMs. Without the LLMs model, ParkingTransformer loses its semantic understanding of the parking task, rendering it unable to generate effective decision-making information. Furthermore, as shown in Table $\mathrm{XI}$, 3D positional encoding proves beneficial for the task. Relying solely on TCA and SCA prevents the vehicle from accurately inferring spatial geometric relationships, which leads to suboptimal performance in narrow parking scenarios. Finally, Table $\mathrm{XII}$ indicates that single-stage training is sufficient to meet the convergence requirements for autonomous parking. Although a two-stage strategy that involves pre-training the LLMs followed by training the ParkingTransformer decoder yields marginally better results, it incurs significantly higher time and computational costs.

\begin{table}[ht]
\centering
\begin{threeparttable}
\begin{minipage}{0.5\textwidth}
\caption{Attention mechanism ablation experiment \label{tab:table9}}
\begin{tabular}{c|c c }
\toprule
Methods                  &Driving Score$\uparrow$  &Success Rate$\uparrow$      \\
\midrule
Full-Attention             & \textbf{61.32}        & \textbf{93.22}                         \\
w/o Temporal-CA            & 59.85                 & 87.43                         \\
w/o Sensor-CA              & 53.19                 & 75.66                         \\
\bottomrule
\end{tabular}
\end{minipage}
\end{threeparttable}
\end{table}

\begin{table}[ht]
\centering
\begin{threeparttable}
\begin{minipage}{0.5\textwidth}
\caption{LLMs ablation experiment \label{tab:table10}}
\begin{tabular}{c|c c }
\toprule
Methods                  &Driving Score$\uparrow$  &Success Rate$\uparrow$      \\
\midrule
Full-LLMs                & \textbf{61.32}            & \textbf{93.22}                           \\
w/o LLMs                 & 54.77                      & 79.59                           \\
\bottomrule
\end{tabular}
\end{minipage}
\end{threeparttable}
\end{table}

\begin{table}[ht]
\centering
\begin{threeparttable}
\begin{minipage}{0.5\textwidth}
\caption{3D PE ablation experiment \label{tab:table11}}
\begin{tabular}{c|c c }
\toprule
Methods                  &Driving Score$\uparrow$  &Success Rate$\uparrow$      \\
\midrule
Full 3D PE              & \textbf{61.32}           & \textbf{93.22}                           \\
w/o 3D PE               & 61.44                 & 94.33                           \\
\bottomrule
\end{tabular}
\end{minipage}
\end{threeparttable}
\end{table}

\begin{table}[ht]
\centering
\begin{threeparttable}
\begin{minipage}{0.5\textwidth}
\caption{Training strategy ablation experiment \label{tab:table12}}
\begin{tabular}{c|c c}
\toprule
Methods                  &Driving Score$\uparrow$  &Success Rate$\uparrow$       \\
\midrule
One-Stage Training       & 61.32                   & 93.22                       \\
Two-Stage Training       &\textbf{61.86}          &\textbf{94.56}              \\
\bottomrule
\end{tabular}
\end{minipage}
\end{threeparttable}
\end{table}

\begin{table}[H]
\centering
\begin{threeparttable}
\begin{minipage}{0.5\textwidth}
\caption{Robustness experiment \label{tab:table13}}
\begin{tabular}{c|c c }
\toprule
Methods                  &Driving Score$\uparrow$    &Success Rate$\uparrow$      \\
\midrule
Camera Crash             & 60.33                     & 87.19                        \\
Image Blur               & 60.85                     & 88.24                         \\
\bottomrule
\end{tabular}
\end{minipage}
\end{threeparttable}
\end{table}

\subsection{ Robustness Testing}

Autonomous parking tasks typically operate within complex environments, making sensor failures an inevitable challenge. We evaluated the system under two specific scenarios: single-camera crash, and single-image blurring. As illustrated in Table $\mathrm{XIII}$, the experimental results indicate that ParkingTransformer has good robustness. This resilience is primarily attributed to the model's retention of historical information and its direct interaction with raw images. Consequently, ParkingTransformer is capable of effectively handling temporary failures of both the camera and the LLMs.

\section{Conclusion}

This paper presents ParkingTransformer, a novel framework that leverages multi-view perception and the scene understanding capabilities of LLMs. Without relying on dense BEV representations, our method combines trajectory queries with the implicit state features of the LLMs to interact directly with historical information and raw sensor data for trajectory planning. We introduce 3D positional encoding to compensate for the limited spatial reasoning capabilities of LLMs and design a fixed-window streaming mechanism to enhance efficiency in long-term temporal processing. Furthermore, the decoder employs a coarse-to-fine strategy to further improve trajectory precision. We conducted extensive closed-loop experiments in CARLA and real-world tests to verify the feasibility and effectiveness of the proposed algorithm.

\section{Acknowledgments}
 The authors gratefully acknowledge the China PostdoctoralScience Foundation under Grants 2025M781569.

\bibliography{new}
\bibliographystyle{IEEEtran}

{\appendix

\begin{enumerate}[
    leftmargin=3em,     
    labelwidth=2em,     
    label=\arabic*),    
    align=right,        
    itemsep=1ex         
]
    \item Target Success Rate (TSR): The success rate is calculated as the ratio of successful trials to the total number of attempts. Specifically, a trial is considered successful only when the ego-vehicle's center is positioned within 0.6 meters of the target center, with a yaw deviation of less than 10 degrees.

    \item Target Failure Rate (TFR): Ego vehicle parked in the parking space, but the distance error exceeds 0.6m, or the yaw error exceeds 10 degrees.

    \item Not Target Rate (NTR): Ego vehicles parked in non-target parking spaces.

    \item Collision Rate (CR): The percentage of collisions that occur during the parking process of ego vehicles to the total number of parking attempts.

    \item Average Position Error (APE): In the case of successful parking of ego vehicles in a parking space, the average distance error between the center of the Ego vehicle and the center of the parking space.

    \item Average Orientation Error (AOE): In the case of successful parking of an ego vehicle in a parking space, the average error between the ego vehicle's yaw angle and the direction of the parking space.

    \item Average Parking Time (AOT): In the case of successful parking of ego vehicles in a parking space, the average time taken from the starting point to the destination.
\end{enumerate}

}


\end{document}